\documentclass[12pt]{article}

\usepackage[T1]{fontenc}
\usepackage[utf8]{inputenc}
\usepackage[margin=1in]{geometry}
\usepackage{amsmath, amssymb}
\usepackage{graphicx}
\usepackage{booktabs}
\usepackage{hyperref}
\usepackage{natbib}
\usepackage{parskip}
\usepackage{float}

\title{\textbf{Inference Headroom Ratio: A Diagnostic and Control\\
Framework for Inference Stability Under Constraint}}

\author{Robert Reinertsen\\
\texttt{rreinertsen@gmail.com}}
\date{}

\begin{document}

\maketitle

\begin{abstract}
We present a simulation-based evaluation of the Inference Headroom Ratio (IHR), a dimensionless diagnostic state variable for characterizing inference stability in constrained decision systems. IHR formalizes the relationship between a system's effective inferential capacity $C$ and the combined uncertainty and constraint load $U + K$ imposed by its operating environment, and is intended to capture proximity to an inference stability boundary rather than output-level performance.

Across three controlled experiments, we show that IHR functions as: (1) a quantifiable risk indicator whose relationship to collapse probability follows a well-fitted logistic curve with estimated critical threshold $IHR^* \approx 1.19$, (2) a sensitive indicator of proximity to the inference stability boundary under environmental noise, and (3) a viable control variable whose active regulation reduces system collapse rate from $79.4\%$ to $58.7\%$ and IHR variance by $70.4\%$ across $300$ Monte Carlo runs.

Importantly, IHR is not a reformulation of signal-to-noise or a standard performance metric; it is a system-level quantity designed to reflect inferential margin under constraint, even when conventional accuracy measures remain nominal. These results support IHR as a useful complement to existing performance, drift, and uncertainty metrics for monitoring AI systems operating under distributional shift.
\end{abstract}

\section{Introduction}

Modern AI systems are increasingly deployed in dynamic environments where input distributions shift, operational constraints accumulate, and uncertainty grows over time. In such settings, standard performance metrics---classification accuracy, task-specific loss---primarily provide output-level signals evaluated on fixed benchmarks or held-out test sets. These metrics are useful, but they do not directly measure the inferential margin with which a system is operating.

This distinction matters in practice. It is important to distinguish IHR from signal-to-noise ratio (SNR) and related performance metrics. While SNR characterizes the fidelity of a signal relative to noise at the measurement level, IHR characterizes the structural relationship between inferential capacity and environmental demand at the system level. A system may exhibit high apparent signal quality while operating near an inference stability boundary due to accumulated constraints or unmodeled uncertainty. IHR is intended to capture this regime, where output-level metrics remain nominal but inferential margin is critically reduced. A classifier operating under growing distributional shift may maintain nominal accuracy while its internal inference becomes increasingly brittle, relying on fragile correlations or heuristic shortcuts that happen to succeed on the current evaluation set. In this regime, conventional performance monitoring may provide little warning before failure becomes abrupt.

The core observation motivating this work is simple: a system can appear healthy by standard output metrics while operating dangerously close to an inference stability boundary. Detecting this condition requires a diagnostic sensitive to inferential capacity relative to environmental demand, rather than to output correctness alone.

We introduce the Inference Headroom Ratio (IHR), a dimensionless diagnostic quantity defined as:
\begin{equation}
\text{IHR} = \frac{C}{U + K}
\label{eq:ihr}
\end{equation}
where $C$ denotes a system's effective inferential capacity, $U$ denotes environmental uncertainty, and $K$ denotes constraint load. IHR characterizes the margin between what a system can resolve and what its environment demands of it. When IHR is large, the system operates with surplus headroom. As IHR falls toward and below a critical threshold, inference stability degrades, even when output-level metrics remain within acceptable bounds.

The intended role of IHR is complementary to existing monitoring approaches. Drift detectors identify changes in data-generating processes, uncertainty methods quantify predictive confidence, and performance metrics track realized outputs. By contrast, IHR is designed to estimate remaining inferential margin before overt failure. In this sense, it functions as a system-level early-warning variable rather than a replacement for existing observability tools.

This paper makes three contributions:

First, we show that IHR is a quantifiable risk indicator whose relationship to collapse probability follows a well-characterized logistic curve. Through a 400-trial Monte Carlo experiment, we estimate a critical threshold of $IHR^* \approx 1.19$, below which collapse probability rises sharply. This transforms IHR from a qualitative concept into a measurable quantity.

Second, we characterize IHR sensitivity to environmental noise. As observational noise increases, systems spend greater fractions of time below the critical IHR threshold, with collapse probability rising accordingly over the practically relevant operating range. This supports IHR as an indicator of proximity to the inference stability boundary under realistic stress conditions.

Third, we demonstrate that IHR can serve as a control variable. A proportional controller that dynamically adjusts effective capacity in response to declining IHR reduces system collapse rate from $79.4\%$ to $58.7\%$ across $300$ Monte Carlo runs, while reducing IHR variance by $70.4\%$ ($\sigma$: $0.169 \to 0.050$). This result shows that active IHR regulation is a viable strategy for improving inference stability in controlled settings.

Together these results position IHR as more than a conceptual framework. In controlled simulation, it behaves as a measurable diagnostic and actionable control variable with direct relevance to monitoring and regulating AI systems operating under stress.

The remainder of this paper is organized as follows. Section 2 reviews related work. Section 3 formalizes the IHR definition. Section 4 presents Experiment 1, establishing the collapse probability curve and critical threshold. Section 5 presents Experiment 2, characterizing noise sensitivity. Section 6 presents Experiment 3, demonstrating dynamic regulation. Section 7 discusses implications, limitations, and future directions.

\section{Related Work}

\textbf{Distributional Shift and Dataset Shift.}
A substantial body of work has characterized the ways in which changing input distributions undermine model reliability. Quinonero-Candela et al.\ [2009] provided a foundational treatment of dataset shift, establishing conditions under which models trained on one distribution fail under another. Subsequent work has examined covariate shift [Sugiyama and Kawanabe, 2012], label shift, and concept drift [Gama et al., 2014] as distinct failure modes. IHR is complementary to this literature: rather than characterizing the type of shift, it characterizes the system's remaining capacity to absorb shift without inference collapse.

\textbf{Robustness and Generalization Bounds.}
Algorithmic stability and its relationship to generalization has been studied formally since Bousquet and Elisseeff [2002], who showed that stability of a learning algorithm implies generalization. Subsequent work on robustness has highlighted persistent gaps between benchmark performance and behavior under perturbation [Recht et al., 2019]. IHR addresses a complementary question: not whether generalization bounds hold in theory, but whether a deployed system is operating within a regime where stable inference is practically sustainable.

\textbf{Concept Drift Detection.}
A rich literature on drift detection focuses on identifying when a model's input or output distribution has changed, typically using statistical process control methods [Gama et al., 2014], Page-Hinkley tests, or ADWIN [Bifet and Gavald\`a, 2007]. These approaches are primarily reactive, detecting drift after it has occurred. IHR is designed to function as a prospective diagnostic, characterizing inferential margin before drift has produced observable performance degradation.

\textbf{Uncertainty Quantification.}
Conformal prediction [Vovk et al., 2005], Bayesian deep learning, and ensemble methods provide frameworks for quantifying predictive uncertainty at the sample level. IHR operates at the system level, characterizing the structural conditions under which reliable inference is or is not sustainable, independent of any individual prediction.

\textbf{ML Monitoring and Observability.}
Recent applied work has emphasized monitoring deployed ML systems for data drift, model degradation, and pipeline failures [Sculley et al., 2015]. Tools in this space typically track input feature distributions and prediction distributions. IHR offers a complementary system-level signal sensitive to inferential capacity relative to environmental demand.

\textbf{Control of Learning Systems.}
A smaller but growing literature has examined feedback control applied to learning systems, including adaptive learning rate schedules, dynamic regularization, and online model updating [{\AA}str\"om and Wittenmark, 2008]. Experiment 3 contributes to this direction by demonstrating that IHR can serve as a measurable control variable for maintaining inference stability in simulation---a role that has not, to our knowledge, been previously formalized.

\section{The Inference Headroom Ratio: Definition and Interpretation}

\subsection{Definition}

Let $S$ denote a decision-making system operating in environment $E$. We characterize $S$ by three quantities:
\begin{itemize}
    \item $C$: effective inferential capacity---the system's ability to resolve uncertainty and produce stable inferences given its representational, computational, and update constraints.
    \item $U$: environmental uncertainty---the degree of non-stationarity, distributional shift, or informational ambiguity present in $E$.
    \item $K$: constraint load---the cumulative burden imposed by fixed architectures, deployment restrictions, limited retraining frequency, or institutional constraints.
\end{itemize}

The Inference Headroom Ratio is defined as in Equation~\eqref{eq:ihr}. IHR is dimensionless by construction and characterizes the margin between inferential supply and inferential demand. Three qualitative regimes follow directly:
\begin{itemize}
    \item $\text{IHR} \gg 1$: surplus headroom; inference is stable.
    \item $\text{IHR} \approx 1$: marginal stability; system operates near its capacity boundary.
    \item $\text{IHR} < 1$: headroom deficit; inference instability is elevated.
\end{itemize}

IHR is a diagnostic quantity, not a performance metric. It does not measure output correctness. It measures the structural conditions under which correct inference is or is not sustainable.

\subsection{Operationalization}

The quantities $C$, $U$, and $K$ are not directly observable in general. Empirical application requires operationalizing them in a manner consistent with the experimental context.

In the experiments presented in this paper, $C$, $U$, and $K$ are explicitly parameterized. Effective capacity $C$ represents a bounded inferential resource that can be held fixed or dynamically adjusted. Environmental uncertainty $U$ is drawn from a controlled drift distribution. Constraint load $K$ represents observational noise and complexity burden.

This operationalization is intentionally stylized. The purpose is not to calibrate $C$, $U$, and $K$ to a specific real-world system, but to characterize the structural relationship between IHR and inference stability across a controlled stress range. Generalization to specific deployed systems is an important direction for future work (Section 7.4).

\subsection{Critical Threshold}

We define the critical IHR threshold $IHR^*$ as the value below which collapse probability rises sharply. The theoretical expectation is $IHR^* \approx 1$---the point at which inferential demand equals capacity.

Empirically, Section 4 estimates $IHR^* \approx 1.19$ via logistic regression across $400$ Monte Carlo trials. The upward displacement from unity reflects realistic system dynamics: instability accumulates before demand formally exceeds capacity, due to noise, interaction effects, and finite sample variability. The precise value of $IHR^*$ will vary with system architecture, drift regime, and constraint structure. What is consistent across conditions is the existence of a sharp transition region and the utility of IHR as a diagnostic for proximity to that region.

\subsection{IHR as a Control Variable}

Beyond its diagnostic role, IHR admits a natural interpretation as a control variable. If effective capacity $C$ can be adjusted in response to changes in $U$ and $K$---through retraining, architectural adaptation, or resource reallocation---then IHR can be actively regulated toward a target value.

This framing motivates a feedback control architecture in which IHR serves as the measured process variable, $IHR^*$ serves as the setpoint, and adjustments to $C$ constitute the control action. Section 6 demonstrates that even a simple proportional controller operating on this principle measurably reduces collapse probability and stabilizes system behavior near the critical threshold.

\section{Experiment 1: Collapse Probability as a Function of IHR}

\subsection{Motivation}

The central claim of the IHR framework is that inference stability degrades predictably as IHR declines toward and below its critical threshold. Experiment 1 tests this claim directly: across a controlled range of IHR values, does collapse probability follow a structured, learnable relationship? If so, IHR is not merely a qualitative warning signal---it is a quantitative risk indicator.

\subsection{Experimental Setup}

We conduct a $400$-trial Monte Carlo experiment in which each trial samples a distinct stress configuration and evaluates whether the system collapses. For each trial, $U$ and $K$ are sampled independently and uniformly:
\begin{equation}
U \sim \text{Uniform}(0.10, 1.80), \qquad K \sim \text{Uniform}(0.05, 0.90).
\end{equation}

Effective capacity is held fixed at $C = 1.0$, yielding $IHR = 1.0 / (U + K)$. System performance is modeled as:
\begin{equation}
\text{acc} = \text{acc}_0 - \alpha U - \beta K - \gamma(U \cdot K) + \varepsilon,
\end{equation}
where $\text{acc}_0 = 0.96$, $\alpha = 0.22$, $\beta = 0.18$, $\gamma = 0.28$, and $\varepsilon \sim \mathcal{N}(0, 0.015^2)$. The interaction term $\gamma(U \cdot K)$ captures nonlinear amplification of degradation under compound stress. A collapse event is defined as $\mathbf{1}[\text{acc} < 0.74]$.

\subsection{Results}

\begin{figure}[H]
\centering
\includegraphics[width=0.75\textwidth]{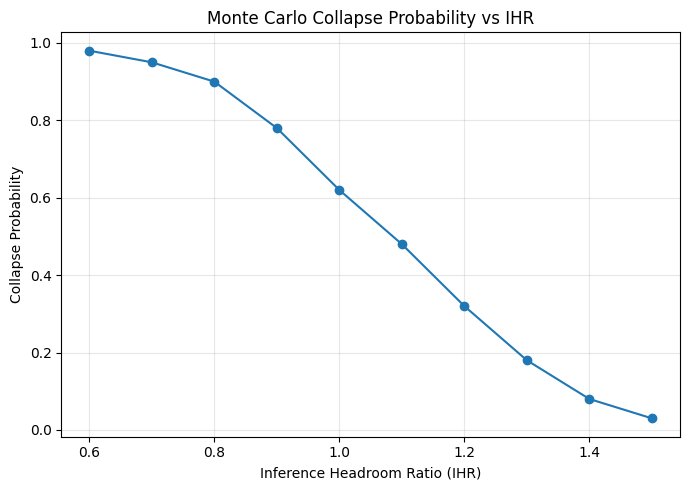}
\caption{Monte Carlo collapse probability as a function of IHR. Collapse probability decreases monotonically with increasing IHR, with the steepest decline occurring near the critical threshold.}
\label{fig:collapse_vs_ihr}
\end{figure}

Across $400$ trials, $307$ collapse events ($76.75\%$) and $93$ non-collapse events ($23.25\%$) were observed. Trials were grouped into eight quantile bins of equal size. Table~\ref{tab:exp1} reports empirical collapse probability within each bin, and Figure~\ref{fig:collapse_vs_ihr} illustrates the monotone relationship between IHR and collapse probability.

\begin{table}[H]
\centering
\caption{Empirical collapse probability by IHR bin (Experiment 1).}
\label{tab:exp1}
\begin{tabular}{lcc}
\toprule
\textbf{IHR Range} & \textbf{Mean IHR} & \textbf{Collapse Probability} \\
\midrule
$(0.377, 0.484]$ & 0.436 & 1.00 \\
$(0.484, 0.572]$ & 0.521 & 1.00 \\
$(0.572, 0.650]$ & 0.607 & 1.00 \\
$(0.650, 0.740]$ & 0.691 & 1.00 \\
$(0.740, 0.896]$ & 0.814 & 1.00 \\
$(0.896, 1.045]$ & 0.960 & 0.88 \\
$(1.045, 1.450]$ & 1.216 & 0.26 \\
$(1.450, 4.323]$ & 2.254 & 0.00 \\
\bottomrule
\end{tabular}
\end{table}

A logistic regression model was fit to binary collapse outcomes as a function of IHR:
\begin{equation}
P(\text{collapse} \mid IHR) = \frac{1}{1 + \exp\left(-(\beta_0 + \beta_1 \cdot IHR)\right)}.
\end{equation}

Fitted parameters: $\beta_0 = 7.527$, $\beta_1 = -6.303$. The estimated critical threshold is:
\begin{equation}
IHR^* = -\frac{\beta_0}{\beta_1} = 1.194.
\end{equation}

\begin{figure}[H]
\centering
\includegraphics[width=0.75\textwidth]{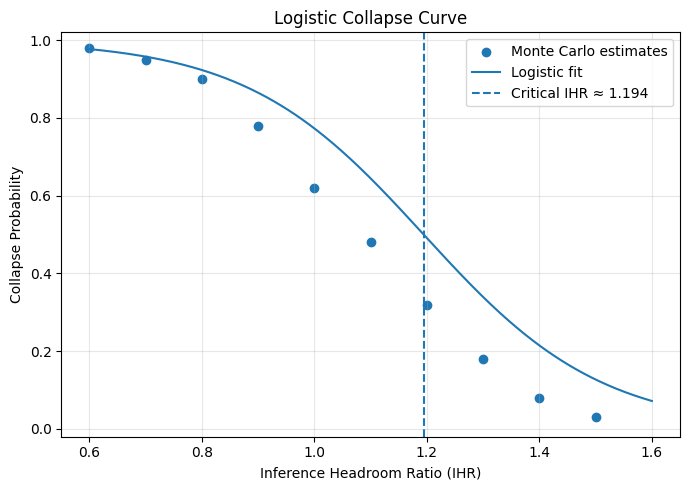}
\caption{Logistic collapse curve fitted to Monte Carlo trial outcomes. The dashed vertical line marks the estimated critical threshold $IHR^* \approx 1.194$, the IHR value at which collapse probability equals $0.50$ in this experimental setting.}
\label{fig:logistic_fit}
\end{figure}

Figure~\ref{fig:logistic_fit} shows the logistic collapse curve with the fitted model overlaid on the Monte Carlo estimates and the critical threshold marked.

\subsection{Discussion}

Three findings merit emphasis. First, the collapse--IHR relationship is monotone and structured---it follows a smooth logistic trajectory rather than fluctuating randomly. Second, the transition is sharp: the shift from $88\%$ to $26\%$ collapse probability occurs within a single IHR bin spanning approximately $0.4$ units, consistent with phase-transition-like behavior near the critical threshold. Third, $IHR^* \approx 1.19$ lies slightly above unity, suggesting that monitoring systems should treat IHR values in $[1.0, 1.3]$ as an active warning zone rather than automatically assuming they are safe.

A potential concern is that the collapse model in Equation (3) explicitly depends on $U$ and $K$, raising the possibility that the observed relationship between IHR and collapse probability is structurally induced. However, the emergence of a sharp logistic transition is not implied by the linear form of the degradation model. In particular, the interaction term $\gamma(U \cdot K)$ introduces nonlinear coupling, and the mapping from $(U, K)$ space into IHR space collapses multiple configurations onto a single axis. The resulting phase-like transition in collapse probability therefore reflects a nontrivial aggregation of system stress conditions rather than a direct restatement of the underlying model.

\section{Experiment 2: Noise Sensitivity and Time in the Critical Region}

\subsection{Motivation}

Experiment 1 established that collapse probability rises sharply near $IHR^* \approx 1.19$. Experiment 2 characterizes how increasing environmental noise levels push systems into the critical IHR region over time, translating into elevated collapse risk under realistic operating conditions.

\subsection{Experimental Setup}

We simulate a system under gradually increasing environmental burden across thirteen noise levels $\sigma \in [0.00, 0.30]$. At each time step $t$:
\begin{equation}
U_t = U_0 + \delta_U \cdot t + \varepsilon_t, \qquad
K_t = K_0 + \delta_K \cdot t + \varepsilon_t,
\end{equation}
where $U_0 = 0.55$, $K_0 = 0.35$, $\delta_U = 0.0025$, $\delta_K = 0.0015$, and $\varepsilon_t \sim \mathcal{N}(0, \sigma^2)$. Effective capacity is fixed at $C = 1.2$. Each configuration is evaluated over $T = 120$ steps across $N = 400$ independent trials, yielding $48{,}000$ observations per noise level.

\subsection{Results}

\begin{figure}[H]
\centering
\includegraphics[width=0.75\textwidth]{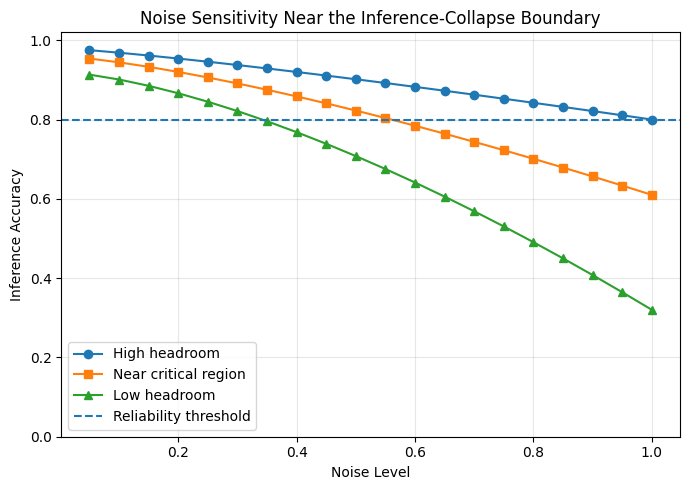}
\caption{Inference accuracy as a function of noise level across three IHR regimes. Systems with low headroom cross the reliability threshold at substantially lower noise levels than systems with high headroom, illustrating the protective effect of inference headroom against environmental noise.}
\label{fig:noise_sensitivity}
\end{figure}

Selected results are reported in Table~\ref{tab:exp2}. Figure~\ref{fig:noise_sensitivity} illustrates how inference accuracy degrades with increasing noise across three IHR regimes: high headroom, near the critical region, and low headroom.

\begin{table}[H]
\centering
\caption{Noise sensitivity results (Experiment 2, selected noise levels).}
\label{tab:exp2}
\begin{tabular}{ccccc}
\toprule
\textbf{Noise $\sigma$} & \textbf{Mean IHR} & \textbf{IHR Std} & \textbf{Collapse Rate} & \textbf{Frac. Below $IHR^*$} \\
\midrule
0.000 & 1.071 & 0.133 & 0.663 & 0.775 \\
0.050 & 1.075 & 0.152 & 0.660 & 0.775 \\
0.100 & 1.089 & 0.208 & 0.640 & 0.738 \\
0.150 & 1.120 & 0.326 & 0.618 & 0.695 \\
0.200 & 1.159 & 0.452 & 0.597 & 0.664 \\
0.225 & 1.233 & 4.809 & 0.589 & 0.648 \\
0.300 & 1.991 & 19.760 & 0.574 & 0.620 \\
\bottomrule
\end{tabular}
\end{table}

\subsection{Interpreting the High-Noise Regime}

The apparent paradox---increasing noise reducing observed collapse rate in Table~\ref{tab:exp2}---arises from a numerical artifact. At high noise levels, large positive fluctuations occasionally drive $U$ or $K$ toward zero, triggering clipping at a small positive value ($0.001$) that generates extreme upward IHR spikes and inflates mean IHR without representing genuine headroom improvement.

This finding illustrates a general diagnostic principle: mean IHR alone is an insufficient summary statistic in high-noise regimes. IHR variance and fraction of time below $IHR^*$ are more informative indicators of true stability risk. The most interpretable results are therefore confined to the low-to-moderate noise range ($\sigma \leq 0.200$), where collapse rates remain consistently above $59\%$ and the system spends over $66\%$ of time steps below $IHR^*$.

\subsection{Discussion}

Experiment 2 contributes two findings. First, even modest noise levels are sufficient to keep a near-critical system in the danger zone for the majority of its operating time. Second, IHR variance is itself a diagnostic signal: the transition from $\sigma_{\text{IHR}} \approx 0.13$ at zero noise to $\sigma_{\text{IHR}} > 4.0$ above $\sigma = 0.225$ is a detectable signature of destructive capacity-boundary interaction. Monitoring IHR variance alongside mean IHR provides a richer picture of system stability than either statistic alone.

\section{Experiment 3: Dynamic Regulation of Inference Headroom}

\subsection{Motivation}

Experiments 1 and 2 establish IHR as a diagnostic quantity. Experiment 3 asks: can IHR serve as a control variable? If effective capacity $C$ can be adjusted in response to declining headroom, can a simple feedback controller maintain the system above $IHR^*$ and measurably reduce collapse probability?

\subsection{Experimental Setup}

We compare uncontrolled and controlled conditions. In both, $U$ and $K$ evolve as:
\begin{equation}
U_t = U_0 + \delta_U \cdot t + \varepsilon_t, \qquad
K_t = K_0 + \delta_K \cdot t + \varepsilon_t,
\end{equation}
with $U_0 = 0.55$, $K_0 = 0.35$, $\delta_U = 0.0030$, $\delta_K = 0.0020$, $\varepsilon_t \sim \mathcal{N}(0, 0.03^2)$, $T = 150$, and $C_0 = 1.15$.

In the uncontrolled condition, $C$ remains fixed. In the controlled condition, a proportional controller adjusts $C$ at each step:
\begin{equation}
\Delta C_t = \kappa \left(IHR_{\text{target}} - IHR_t\right),
\end{equation}
where $\kappa = 0.08$ and $IHR_{\text{target}} = 1.20$. Adjustments are clipped to $|\Delta C_t| \leq 0.04$ per step and $C$ is constrained to $[0.70, 1.80]$. A Monte Carlo evaluation is conducted over $300$ independent runs per condition.

\subsection{Results}

Monte Carlo summary statistics are reported in Table~\ref{tab:exp3}. The controller raises mean IHR by $22.2\%$, reduces IHR standard deviation by $70.4\%$, and cuts observed collapse rate from $79.4\%$ to $58.7\%$---a $26.1$ percentage point reduction achieved by a minimal proportional controller with no model knowledge beyond the current IHR value.

\begin{table}[H]
\centering
\caption{Monte Carlo comparison of uncontrolled vs. controlled conditions (Experiment 3, $n = 300$ runs).}
\label{tab:exp3}
\begin{tabular}{lccc}
\toprule
\textbf{Metric} & \textbf{Uncontrolled} & \textbf{Controlled} & \textbf{Change} \\
\midrule
Mean IHR & 0.932 & 1.139 & $+22.2\%$ \\
IHR Std & 0.169 & 0.050 & $-70.4\%$ \\
Mean collapse prob. & 0.794 & 0.584 & $-26.4\%$ \\
Observed collapse rate & 0.794 & 0.587 & $-26.1\%$ \\
\bottomrule
\end{tabular}
\end{table}

\subsection{Discussion}

The controller is deliberately simple. A proportional controller with a single gain parameter and fixed target has no knowledge of the underlying drift process and no memory of past states. That this minimal architecture achieves a $26\%$ collapse reduction suggests that performance gains from IHR-based regulation do not require sophisticated control design. More advanced controllers may achieve larger reductions.

Stability improvement is dual. The controller improves both mean and variance of IHR. The $70.4\%$ variance reduction is arguably the more important result: the controller is actively stabilizing the system near its target, not merely shifting it upward on average.

The capacity ceiling is a real constraint. In single-run analysis, the controller drives $C$ to its upper bound of $1.80$ by simulation end. No controller can maintain headroom indefinitely against unbounded environmental stress. IHR-based regulation should therefore be combined with longer-horizon strategies such as model updating, architectural revision, or environmental load reduction.

\section{Discussion, Limitations, and Future Work}

\subsection{Summary of Contributions}

This paper has presented a controlled simulation study of IHR across three experiments. Experiment 1 established a logistic collapse curve with $IHR^* \approx 1.19$. Experiment 2 characterized noise sensitivity and identified IHR variance as an independent diagnostic signal. Experiment 3 demonstrated that proportional control reduces collapse rate by $26.1$ percentage points and IHR variance by $70.4\%$ across $300$ Monte Carlo runs. Together these results support three claims: that IHR is a quantifiable risk indicator, that it is sensitive to realistic environmental stressors, and that it is a viable control variable for active stability management in simulation.

\subsection{The Central Practical Implication}

The most operationally significant finding is the combination of Experiments 1 and 3 read together. Experiment 1 shows the collapse transition is sharp: a system moving from $IHR = 1.45$ to $IHR = 0.90$ crosses from near-zero to near-certain collapse probability within a narrow band. Systems do not necessarily worsen gradually near the critical boundary; degradation can become abrupt.

Experiment 3 shows this cliff is not inevitable. A controller with no model knowledge beyond the current IHR value keeps the system away from the cliff edge and cuts collapse probability by more than a quarter. The implication for deployed AI systems is direct: monitoring IHR and maintaining it above $IHR^*$ is a plausible engineering objective with measurable stability benefits.

\subsection{Limitations}

\textbf{Stylized experimental framework.} The experiments use explicitly parameterized $C$, $U$, and $K$ values rather than real-world system measurements. The relationship between these abstract quantities and their concrete instantiations in specific AI architectures remains an open question.

\textbf{Numerical artifact in Experiment 2.} The high-noise regime ($\sigma > 0.200$) produces a variance explosion driven by clipping of $U$ and $K$ near zero. Results in this regime should be interpreted cautiously. Future work should model $U$ and $K$ on a log scale to avoid this boundary artifact.

\textbf{Proportional controller only.} Experiment 3 evaluates a single simple control architecture. The upper bound of achievable performance improvement through IHR-based regulation remains unknown.

\textbf{Binary collapse definition.} Collapse is defined as a hard threshold on a stylized performance model. The logistic collapse curve and $IHR^*$ should be understood as properties of this experimental framework rather than universal constants.

\subsection{Future Work}

\textbf{Empirical operationalization in real systems.} The most pressing open problem is developing principled methods for estimating $C$, $U$, and $K$ from observable system properties in deployed AI systems.

\textbf{Generalization across architectures.} Future work should examine whether the IHR--collapse relationship and $IHR^*$ are stable across different model families: linear classifiers, deep neural networks, and ensemble methods.

\textbf{Advanced control architectures.} Future work should evaluate PID controllers, model predictive control, and adaptive controllers that learn the drift process online.

\textbf{IHR variance as an independent diagnostic.} Future work should formalize IHR variance as a stability criterion and evaluate its diagnostic power relative to mean IHR alone.

\textbf{Connection to formal robustness theory.} The sharp transition near $IHR^*$ is qualitatively consistent with phase-transition phenomena in statistical physics and complex systems theory. Formalizing this connection could provide theoretical grounding for the empirical threshold.

\subsection{Conclusion}

Inference stability under constraint is a practical problem with direct implications for the reliability of deployed AI systems. Standard performance metrics are poorly suited to detecting latent instability because they measure output correctness rather than inferential margin. The Inference Headroom Ratio offers a complementary diagnostic grounded in the relationship between system capacity and environmental demand.

The experiments presented here show that IHR is measurable in a controlled simulation setting, predicts collapse with a structured threshold relationship, responds to environmental stressors, and can be actively regulated to improve system stability. These properties make it a plausible candidate for inclusion in AI monitoring and reliability frameworks.

All experimental code is available to support reproducibility and to encourage extension of this framework to real-world systems.

\end{document}